\pdfoutput=1

\documentclass[11pt]{article}

\usepackage{acl}

\usepackage{times}
\usepackage{latexsym}

\usepackage[T1]{fontenc}

\usepackage[utf8]{inputenc}

\usepackage{hyperref}       
\usepackage{url}            
\usepackage{booktabs}       
\usepackage{amsfonts}       
\usepackage{nicefrac}       
\usepackage{microtype}      
\usepackage{xcolor}         
\usepackage{microtype}
\usepackage{mathrsfs}
\usepackage{amsmath}
\usepackage{caption}
\usepackage{subcaption}
\usepackage{amssymb}
\usepackage{algorithm} 
\usepackage{algpseudocode} 
\usepackage{soul}
\usepackage{xcolor}
\usepackage{ulem}
\usepackage{mwe}
\usepackage{wrapfig}
\usepackage{multirow}
\usepackage{colortbl,tabularx}
\usepackage{placeins}
\usepackage{subcaption,siunitx,booktabs}
\usepackage{adjustbox}
\newcolumntype{L}[1]{>{\raggedright\arraybackslash}m{#1}}
\newcolumntype{C}[1]{>{\centering\arraybackslash}m{#1}}
\newcolumntype{R}[1]{>{\raggedleft\arraybackslash}m{#1}}

\usepackage{tablefootnote}

\newcommand{\mname}{\textsc{ERNIE-Sparse}\xspace}
\usepackage{algorithm}
\usepackage{algorithmicx}
\algnewcommand\algorithmicinput{\textbf{Input:}}
\algnewcommand\algorithmicoutput{\textbf{Output:}}
\algnewcommand\algorithmicparameter{\textbf{Parameters:}}
\algnewcommand\INPUT{\item[\algorithmicinput]}
\algnewcommand\OUTPUT{\item[\algorithmicoutput]}
\algnewcommand\PARAMETER{\item[\algorithmicparameter]}

\definecolor{light_blue}{RGB}{210,225,240}
\definecolor{dark_blue}{RGB}{158,190,216}
\definecolor{green}{RGB}{210,237,202}
\title{\mname: Learning Hierarchical Efficient Transformer Through Regularized Self-Attention}

\author{ \textbf{Yang Liu\textsuperscript{1, 2}\thanks{\textsuperscript{\dag} The work was done when Yang Liu was doing internship at Baidu. }, Jiaxiang Liu\textsuperscript{1}\thanks{\llap{}\:\:\:Corresponding author. }, Li Chen\textsuperscript{2}, Yuxiang Lu\textsuperscript{1}, Shikun Feng\textsuperscript{1}} \\
 \textbf{Zhida Feng\textsuperscript{1,2}, Yu Sun\textsuperscript{1}, Hao Tian\textsuperscript{1}, Hua Wu\textsuperscript{1} and Haifeng Wang\textsuperscript{1}} \\
	\textsuperscript{1}Baidu Inc.; \textsuperscript{2}Wuhan University of Science and Technology \\
	\{\tt liuyang148, liujiaxiang, luyuxiang, \\
	\tt fengshikun01, fengzhida, sunyu02, tianhao,  \\
	\tt wu\_hua, wanghaifeng\}@baidu.com \\
	\tt chenli@wust.edu.cn}

\begin{document}
\maketitle
\begin{abstract}
Sparse Transformer has recently attracted a lot of attention since the ability for reducing the quadratic dependency on the sequence length. We argue that two factors, \textit{information bottleneck sensitivity} and \textit{inconsistency between different attention topologies}, could affect the performance of the Sparse Transformer. This paper proposes a well-designed model  named \mname. It consists of two distinctive parts: (i) \textbf{H}ierarchical \textbf{S}parse \textbf{T}ransformer (\textbf{HST}) to sequentially unify local and global information. (ii) \textbf{S}elf-\textbf{A}ttention \textbf{R}egularization (\textbf{SAR}) method, a novel regularization designed to minimize the distance for transformers with different attention topologies. To evaluate the effectiveness of \mname, we perform extensive evaluations. Firstly, we perform experiments on a multi-modal long sequence modeling task benchmark, Long Range Arena (LRA). Experimental results demonstrate that \mname significantly outperforms a variety of strong baseline methods including the dense attention and other efficient sparse attention methods and achieves improvements by 2.77\% (57.78\% vs. 55.01\%). Secondly, to further show the effectiveness of our method, we pretrain \mname and verified it on 3 text classification and 2 QA downstream tasks, achieve improvements on classification benchmark by 0.83\% (92.46\% vs. 91.63\%), on QA benchmark by 3.24\% (74.67\% vs. 71.43\%). Experimental results continue to demonstrate its superior performance. 
\end{abstract}


\section{Introduction}
\noindent Transformer \citep{vaswani2017attention} architecture is a key component for many pretrained language models such as BERT \citep{devlin2018bert}, RoBERTa \cite{liu2019roberta}, XLNet \citep{yang2019xlnet}, ERNIE \citep{sun2020ernie}, ALBERT \citep{albert}, ELECTRA\citep{electra}, T5 \citep{t5}. Self-attention is one of the most important modules in transformer. It eliminates the sequential dependency constraints of recurrent neural networks by introducing interactions between each token pair to capture contextual information. However, the self-attention’s computational complexity and memory cost grow quadratically with sequence length, which comes with complexity $O(N^2)$ for processing contexts of $N$ inputs. 

One way to optimize self-attention complexity is introducing sparsity into attention layers \citep{child2019generating,qiu2019blockwise,beltagy2020longformer} by having each token attend to only a subset of tokens in the whole sequence. Recent sparse-attention works \citep{child2019generating,beltagy2020longformer,ainslie2020etc,zaheer2020bigbird} introduce global tokens that can attend to the whole sequence. Those global tokens are used as a form of memory to strengthen global information. While this method reduces the complexity of full self-attention, there are two issues with Sparse Transformer that affect performance.

The first issue is \textit{information bottleneck sensitivity} as shown in Figure~\ref{fig:mainfig}. Information bottleneck is a phenomenon caused by the low number of global tokens — the model had to encapsulate the entire input sequence into those global tokens. If the size of the information bottleneck becomes smaller, performance can suffer. The second issue is \textit{inconsistency between different attention topologies}. Normally, a sparse pattern is pre-defined and fixed in practical training. Due to that sparse pattern splitting the input into several blocks and interaction for input is partially connected, minor changes, e.g., shifting the attention input by a few positions, will lead to the inconsistency in the attention topology, as shown in Figure~\ref{fig:saor}. Compared to the effect for vanilla attention, the sparse attention mechanism will amplify the impact of the minor change.

To resolve the aforementioned issues, we propose a well-designed efficient model \mname. Our method proposes two major techniques including \textbf{H}ierarchical \textbf{S}parse \textbf{T}ransformer (\textbf{HST}) and \textbf{S}elf-\textbf{A}ttention \textbf{R}egularization (\textbf{SAR}). The first technique HST is designed for \textit{information bottleneck sensitivity} problem based on hierarchical attention mechanism. HST introduces special tokens to represent local information and global information in two sequential steps, providing an additional way for modeling global information. The second technique SAR, is designed for \textit{inconsistency between different attention topologies}. We introduce a training regularization strategy to force models with different sparse attention topologies to have consistent outputs. 


To evaluate \mname's ability to model long documents, we first perform extensive evaluations on a long sequence modeling task benchmark \citep{tay2020long}: Long Range Arena (LRA). Experiment results demonstrate that our model significantly outperforms existing methods and provides a new robust solution to the long range sequence modeling. Our proposed method achieves improvements by average 2.77 points (57.78\% vs. 55.01\%) on five long sequence tasks of LRA. We also conduct experiments on 3 long document classification and 2 question answering tasks using pretraining and finetuning paradigm, further shows the effectiveness of \mname. Extensive experiments on those 10 tasks demonstrate the effectiveness of our approach in both cold start and pretrain-then-finetune paradigm.

\section{Related Work}

\textbf{Sparse Transformer}
Sparse attention is widely adopted to solve the long range sequence modeling problem. A simple version is that split the sequence into blocks and perform attention only within block  \citep{qiu2019blockwise,liu-local-transformer,image-transformer}. This mechanism is also called local attention because only local tokens within the block can attend to each other. To improve the connection between tokens, global attention is introduced by \citep{child2019generating,beltagy2020longformer,ainslie2020etc,zaheer2020bigbird}. The global attention mechanism mainly relies on specifying some tokens as global tokens. Those global tokens are used as a form of memory to strengthen global information. However, we observed that the mechanism of global and local attention in sparse attention would cause an information bottleneck, which would affect the information flow between local tokens. This phenomenon was also observed in paper \citep{alon2021on} for Graph Neural Network  \citep{gori2005new}. To this end, we propose to use hierarchical mechanism which will be discussed below.

\textbf{Hierarchical Transformer} Hierarchical learning has been suggested by many researchers and it has been empirically shown to be effective in numerous diverse tasks of natural language processing \citep{zhang2019hibert,liu2019hierarchical,rohde2021hierarchical,wu2021hi}. In this paper, we propose to apply a hierarchical mechanism in Sparse Transformer and provide a new perspective from information flow to describe its beneficiary for Sparse Transformer.

\textbf{Regularization Method} 
Although state-of-the-art deep neural networks have achieved high performance on various NLP tasks, the architectures used for these tasks have been shown to be unstable to small modifications of input texts \citep{DBLP:conf/acl/EbrahimiRLD18,DBLP:conf/aaai/JinJZS20,DBLP:journals/corr/abs-2003-04985}. Adversarial training \citep{Szegedy,fgsm,vat,alum} is proposed to to act as a regularization method to improve model performance by constructing adversarial input. However, current regularization methods are mostly designed for dense models and don't take into account the characteristic of sparse model. Recently, \citep{liang2021rdrop} proposed a simple regularization strategy which forces the output distributions of different sub models generated by dropout to be consistent with each other. Inspired by this work, we propose a regularization method built upon sparse attention topology.

\section{Method}
In this section, we first revisit the mechanism of Sparse Transformer in Section \ref{section:revisit}, and describe the two components of sparse attention (i.e. global attention and local attention). In addition, we provide a new perspective for understanding Sparse Transformers from information flow. At the end of Section \ref{section:revisit}, we discuss the problems existing in the mechanism of the Sparse Transformer. Next, we introduce the motivation, formulation and benefits of the HST in Section \ref{section:hst}. Finally, a regularization method, SAR, is illustrated in Section \ref{section:saor}. 

\subsection{Revisiting Sparse Transformer}
\label{section:revisit}

Given a sequence of input $x=(t_1, ... , t_n)$ of length $n$, dense attention can be formulated as: 

\begin{equation*}
\begin{aligned}
\label{equ:qkv}
\left(\begin{array}{l}
\mathbf{Q} \\
\mathbf{K} \\
\mathbf{V}
\end{array}\right)=\mathbf{H}\left(\begin{array}{c}
\mathbf{W}_{\mathbf{q}} \\
\mathbf{W}_{\mathbf{k}} \\
\mathbf{W}_{\mathrm{v}}
\end{array}\right) +\left(\begin{array}{l}
\mathbf{b}_{\mathbf{q}} \\
\mathbf{b}_{\mathrm{k}} \\
\mathbf{b}_{\mathrm{v}}
\end{array}\right), \\
\operatorname{Attention}=\operatorname{Softmax}\left(\mathbf{Q} \mathbf{K}^{T}\right) \mathbf{V},
\end{aligned}
\end{equation*}
\noindent where $\mathbf{Q},\mathbf{K},\mathbf{V} \in \mathbb{R}^{n\times d} $ is linearly mapped from $\mathbf{H}\in\mathbb{R}^{n \times d}$, $d$ is the size of the hidden vector, $\textbf{H}=(h^l_1, ..., h^l_n)$ for layer $l\in{[1, L]}$, $L$ is the number of model layers, $\mathbf{W}_{\{\mathbf{q, k, v}\}}$ are mapping weights, and $\mathbf{b}_{\{\mathbf{q, k, v}\}}$ are bias terms. For simplicity, the scale factor for $\mathbf{Q}$ is omitted in the equation. Computation complexity for dense transformer is $O(n^2)$. In dense attention, tokens are connected to each other while in sparse attention, tokens are only partially connected.

As shown in Figure \ref{fig:mainfig} (a), sparse attention mainly consists of global attention and local attention \citep{liu-local-transformer,image-transformer,child2019generating,beltagy2020longformer,zaheer2020bigbird}. In this Figure, $\mathbf{Q}$, $\mathbf{K}$, $\mathbf{V}$ are divided into 4 blocks respectively, ($t_1$, $t_2\sim t_3$, $t_4\sim t_5$, $t_6\sim t_7$). The light blue part denotes the global attention meaning the global tokens from this part can attend to all other tokens. The dark blue part expresses local attention denoting that the local tokens can only attend to the tokens within the same block. It can be seen from Figure \ref{fig:mainfig} (b) that there is a bottleneck in the information flow of Sparse Transformer. For sparse attention mechanism, tokens in different attention blocks within the same layer are invisible to each other. For example, $t_3$ can't attend to $t_4$ and vice versa. For these tokens, the only way to attend to each other is through global tokens as relay nodes. As shown in Figure \ref{fig:mainfig} (b), the information of $t_3$ and $t_4$ first flows to $t_1$ at layer $l$, and then distributes back to $t_3$ and $t_4$ at layer $l+1$. This mechanism causes the current layer local token information to be carried by only a few global tokens leading to a bottleneck in the flow of information. According to our observation in section \ref{section:bottleneck}, with the information bottleneck size decreases, the performance has a downward trend.

Another problem caused by sparse attention mechanism is inconsistency between the sparse attention topology change. As shown in Figure \ref{fig:saor}, we plot the attention topology of layer $l$ for $\left[t_1, t_2, \cdots, t_8\right]$ in left figure, and for $\left[t_2, \cdots, t_8, t_1\right]$ in right. We see that even only $t_1$ is shifted by one place, the topology pattern changes much, where dense attention will be less affected as it has a fully connected attention topology. To this end, we propose HST and SAR and will further describe them below. 


\subsection{HST: Hierarchical Sparse Transformer} 
\label{section:hst}
\begin{figure*}[tbp]
	\centering
		\includegraphics[width=1.0\textwidth]{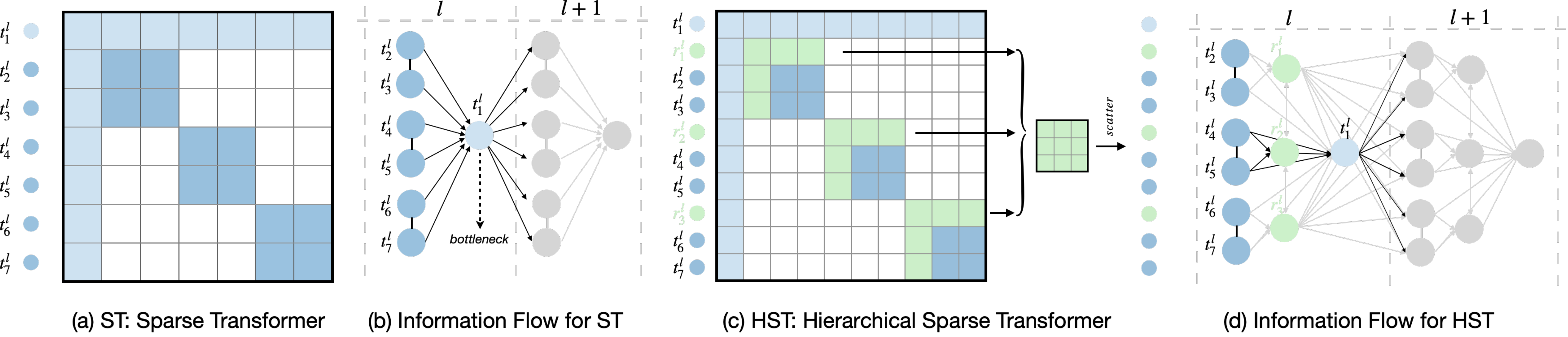}
	\caption{\textbf{The comparison between Sparse Transformer (ST) and Hierarchical Sparse Transformer (HST)}. \textbf{(a)} Sparse Transformer mainly consists of \colorbox{light_blue}{global attention} and \colorbox{dark_blue}{local attention}.
	\textbf{(b)} The bottleneck that existed in ST: all the sequence information is compressed into a fixed size vector. \textbf{(c)} In HST, \colorbox{green}{representative tokens} are inserted into local attention for hierarchical attention. \textbf{(d)} Information flow demonstration for HST: interaction between representative nodes can increase the path of global information interaction.}
	\label{fig:mainfig}
\end{figure*}


HST mainly contains three key elements, i.e. representative tokens insertion, hierarchical attention and the usage for those representative tokens in the last layer for downstream task. 

As aforementioned, the input sample consists of $n$ tokens $(t_1, t_2, ..., t_n)$. We set $g$ as the number of global tokens and $w$ as the block size meaning the number of tokens for each local attention. We propose to insert $m$ representative tokens into input sequence, where $m=(n-g) / w$. Those tokens are inserted to the start position of each block. Similar to BERT \citep{devlin2018bert}, we use $\operatorname{[CLS]}$ for those representative tokens. Thus the encoder input is as follows:
\begin{equation}
\scalebox{0.9}{
\begin{math}
\begin{aligned}
\mathbf{H}^{0} &= 
\underbrace{\mathbf{E}(t_1); \cdots;\mathbf{E}(t_g);}_{global} \\
& \cdots,\underbrace{\mathbf{E}(r_i); \mathbf{E}(t_{g+w(i-1)+1}) ;\cdots;\mathbf{E}(t_{g+wi})}_{i-th \  local},\cdots
\end{aligned}
\end{math}
}
\end{equation}
where $i=\left[1, 2, \cdots, m\right]$, $\mathbf{E}(t) \in \mathbb{R}^d$ is the embedding lookup for token $t$. $\mathbf{E}(r)\in \mathbb{R}^d$ is the representation token embedding. Then we use Sparse Transformer to encode the sequence as follows: 
\begin{equation*}
\begin{aligned}
\mathbf{H}^{l}_s &=\operatorname{SparseTransformer}\left(\left[\mathbf{H}^{l-1}\right]\right),
\end{aligned}
\end{equation*}
where $\mathbf{H}^{l}_s \in \mathbb{R}^{n \times d}$ is the hidden output from one layer Sparse Transformer. To better interact globally, representative tokens are extracted from $\mathbf{H}^{l}_s$ for dense attention as shown in Figure \ref{fig:mainfig} (c). Formally,  the hierarchical attention is calculated by:
\begin{equation*}
\begin{aligned}
\mathbf{R}^{l}_s &=\left(r^{l}_{1} \cdots r^{l}_{m}\right), \\
\mathbf{R}^{l} &=\operatorname{Attention}\left(\mathbf{R}^{l}_s\right),
\end{aligned}
\end{equation*}
where $\mathbf{R}^l_s\in \mathbb{R}^{m\times d}$ is the matrix of representative token's hidden states for layer $l$,  $r^{l}_{1} \cdots r^{l}_{m}$ are extracted from $\mathbf{H}^{l}_s$. $\operatorname{Attention}$ is the dense attention described in \ref{section:revisit}. After dense attention, these representative token hidden vectors in $\mathbf{R}^l$ are distributed back to $\mathbf{H}^{l}_s$ so we get the final hidden vectors $\mathbf{H}^l$ of $l$-th layer. The whole process can be seen more clearly in Figure \ref{fig:mainfig} (c), after hierarchical attention (green matrix), the green dots (denoting representative tokens) are distributed back to the list of tokens. 
The information flow of the hierarchical attention is shown in Figure \ref{fig:mainfig} (d) showing that richer global information interaction path are created. For example, $t_3$ and $t_5$ can complete an interaction by representative nodes in addition to global tokens. 



Note that the $\operatorname{Attention}$ module will introduce additional weights $\mathbf{W}_\mathbf{q}$, $\mathbf{W}_\mathbf{k}$ and $\mathbf{W}_\mathbf{v}$ mentioned in Section \ref{section:revisit}, so how to initialize these three weights needs to be discussed. One option is to randomly initialize these three weights, or we can use the weight of the $\operatorname{SparseTransformer}$ to warm start, or we should also consider whether to share the weights in $\operatorname{Attention}$ with $\operatorname{SparseTransformer}$. Those details will be discussed in Experiment section.

For the downstream task, it is efficient to use the representative tokens as follow:
\begin{equation*}
\begin{aligned}
\mathbf{O}^{L} &=\operatorname{Pooling}\left(\mathbf{R}^{L}\right), \\
\mathcal{P}\left(y \mid x\right)&=\operatorname{Softmax}(\mathbf{O}^{L}\mathbf{W}_o),
\end{aligned}
\end{equation*}
where $\operatorname{Pooling}$ can be $\operatorname{MAX}$, $\operatorname{MEAN}$, $\mathbf{O}^L \in \mathbb{R}^d$, $\mathbf{W}_o \in \mathbb{R}^{d\times c}$, $c$ is the downstream task class number. $y$ is the downstream task label, $\mathcal{P}\left(y \mid x\right)$ denotes the predicted probability. Note that $\operatorname{Pooling}$, can be replaced as $\operatorname{CLS}_g$, where $\operatorname{CLS}_g$ is the first global token. 

\subsection{SAR: Self-Attention Regularization} 
\label{section:saor}
\begin{figure*}[t]
	\centering
		\includegraphics[width=1.0\textwidth]{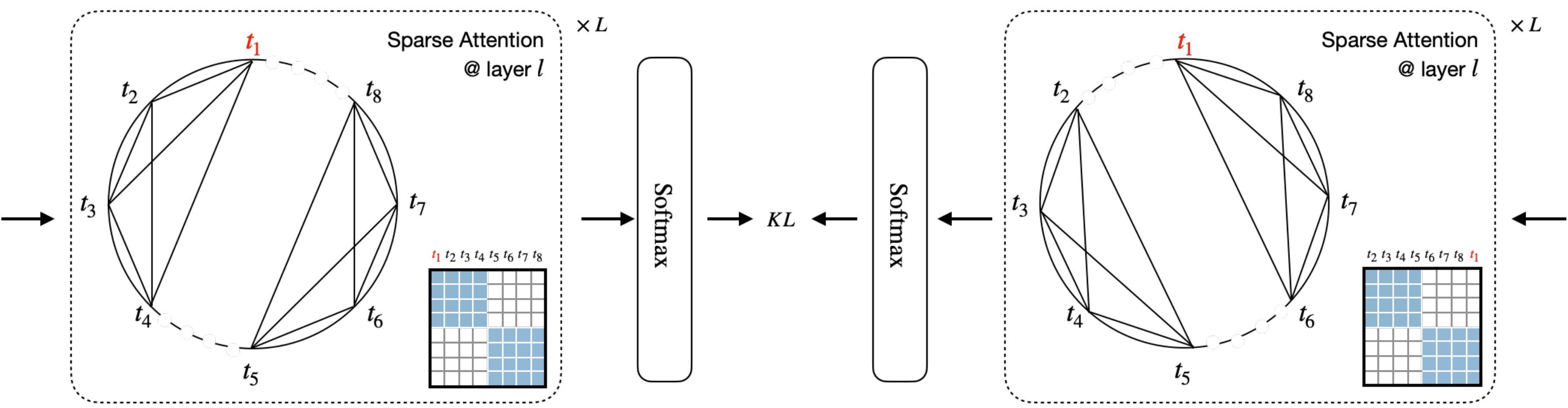}
	\caption{\textbf{The overall framework of our proposed SAR}. The procedure of SAR that regularizes the model outputs of transformers with different attention topologies.}
	\label{fig:saor}
	
\end{figure*}

Different with dense attention, sparse attention doesn't guarantee interaction between all tokens as the design of sparse attention is to allow tokens can only attend to some of other tokens in one layer. This leads to a problem of inconsistency: even shifting by one place to attention input will cause much sparse topology change, as shown in Figure \ref{fig:saor} and discussed in detail in Section \ref{section:revisit}. To this end, we introduce a regularization method to leverage the outputs of Sparse Transformer.


Our method is designed based on the motivation that for the same input, the outputs of transformers with different attention topologies should be consistent. As shown in Figure \ref{fig:saor}, the left shows a $L$ layers Sparse Transformer with the default sparse attention topology, and the right shows a transformer with a shifted attention topology. Concretely, given the input data $x$ at each training step, we feed $x$ to go through the forward pass of the network twice, one with default sparse attention topology, 
and the other with the shifted attention topology. 
Therefore, we can obtain two distributions of the model predictions, denoted $\mathcal{P}_{1}\left(y \mid x\right)$ and $\mathcal{P}_{2}\left(y \mid x\right)$. Thus the distributions of $\mathcal{P}_{1}\left(y \mid x\right)$ and $\mathcal{P}_{2}\left(y \mid x\right)$ are different for the same input data pair $(x, y)$. Then at this training step, in order for those two distributions to be close to each other, our SAR method tries to minimize the bidirectional Kullback-Leibler (KL) divergence between these two output distributions. Formally speaking,
\begin{equation*}
\label{equ:kl}
\begin{aligned}
\mathcal{L}_{S A R}=& \frac{1}{2}\left[\mathcal{D}_{K L}\left(\mathcal{P}_{1}(y \mid x) \| \mathcal{P}_{2}\left(y \mid x\right)\right)\right.\\
&\left.+\mathcal{D}_{K L}\left(\mathcal{P}_{2}\left(y \mid x\right) \| \mathcal{P}_{1}(y \mid x)\right)\right]
\end{aligned}
\end{equation*}
\noindent Assume that the learning objective of target task is negative log-likelihood, the loss of the two forward passes is:
\begin{equation*}
\begin{aligned}
\label{equ:nll}
\mathcal{L}_{N L L}=-\log \mathcal{P}_{1}\left(y \mid x\right)-\log \mathcal{P}_{2}\left(y \mid x\right),
 \end{aligned}
\end{equation*}
as a result, the total loss should be :
\begin{equation*}
\begin{aligned}
\label{equ:total}
\mathcal{L}=\mathcal{L}_{NLL} + \alpha \mathcal{L}_{SAR},
 \end{aligned}
\end{equation*}
where $\alpha$ is the loss coefficient of $\mathcal{L}_{SAR}$.
Our method naturally makes transformer outputs of different attention topologies close to each other.


\section{Experiments}
\label{exp}

To evaluate our approach and show its performance, we first conduct experiments on a long context sequence modeling benchmark, LRA, which is consisted of 5 multi-modal tasks including logical inference, natural language and image tasks. LRA is a benchmark for cold start models and does not require the model to be pre-trained, so LRA is a suitable benchmark for testing the model structure designs. To further evaluate the ability of \mname in the pretrain-then-finetune paradigm, we also follow \citep{beltagy2020longformer} and \citep{zaheer2020bigbird} to pretrain \mname  and test the pretrained \mname on 3 natural language classification and 2 question answering tasks.

\subsection{Long-Context Sequence Modeling}
\label{section:main_lra}
\begin{table*}[t]
    \centering
    \small
\begin{center}
\resizebox{0.98\textwidth}{!}
{

\begin{tabular}{l|ccccc|c}
  \toprule
  \toprule
    \textbf{Models} &  \textbf{ListOps} & \textbf{Text}  & \textbf{Retrieval} & \textbf{Image} & \textbf{Pathfinder} & \textbf{Avg} \\ 
\midrule
    Local Attention                             & 15.82             & 52.98                 & 53.39 
                                                & 41.46             & 66.63                 & 46.06 \\
    Linear Trans. \citep{linear_trans}           & 16.13             & \textbf{65.90}                 & 53.09 
                                                & 42.34             & 75.30                 & 50.55 \\
    Reformer \citep{reformer}                    & \underline{37.27} & 56.10                 & 53.40 
                                                & 38.07             & 68.50                 & 50.67 \\
    Sparse Trans. \citep{child2019generating}    & 17.07             & 63.58                 & \underline{59.59} 
                                                & \underline{44.24} & 71.71                 & 51.24 \\
    Sinkhorn Trans.\citep{sinkhorn}              & 33.67             & 61.20                 & 53.83 
                                                & 41.23             & 67.45                 & 51.39 \\
    Linformer \citep{linformer}                  & 35.70             & 53.94                 & 52.27 
                                                & 38.56             & 76.34                 & 51.36 \\
    Performer \citep{performer}                  & 18.01             & \underline{65.40}        & 53.82 
                                                & 42.77             & \underline{77.05}     & 51.41 \\
    Synthesizer \citep{synthesizer}              & 36.99             & 61.68                 & 54.67            
                                                & 41.61             & 69.45                 & 52.88 \\
    Longformer \citep{beltagy2020longformer}     & 35.63             & 62.85                 & 56.89            
                                                & 42.22             & 69.71                 & 53.46 \\
    Transformer \citep{vaswani2017attention}     & 36.37             & 64.27                 & 57.46            
                                                & 42.44             & 71.40                 & 54.39 \\
    BigBird \citep{zaheer2020bigbird}            & 36.05             & 64.02                 & 59.29            
                                                & 40.83             & 74.87                 & \underline{55.01} \\
\midrule
    \mname                                & \textbf{37.75}    & 64.47                 & \textbf{62.64}   
                                                & \textbf{45.28}    & \textbf{78.77}        & \textbf{57.78} \\
\bottomrule
\end{tabular}
} 
\end{center}
\caption{Experimental results on the long range arena (LRA) benchmark. The highest acc for each dataset is highlighted in bold and the second place is underlined.}

\setlength{\belowcaptionskip}{-4cm}

\label{lra_experimental_results}
\end{table*}

We first evaluate the effectiveness and efficiency of \mname on the LRA benchmark recently introduced by \citep{lra}, which is designed to evaluate efficient transformer models under the long-context scenario. They collect five tasks in this benchmark which are ListOps \citep{DBLP:conf/naacl/NangiaB18}, byte-level text classification \citep{DBLP:conf/acl/MaasDPHNP11}, byte-level document retrieval \citep{DBLP:journals/lre/RadevMQA13}, image classification on sequences of pixels \citep{krizhevsky2009learning} and Pathfinder \citep{DBLP:conf/nips/LinsleyKVWS18}. 
 This benchmark consists of sequences ranging from 1K to 16K tokens, encompassing a wide range of data types and modalities such as text, natural, synthetic images, and mathematical expressions requiring similarity, structural, and visual-spatial reasoning. We run each experiment five times with different random seeds and report the average accuracy.

The result of \mname on the LRA tasks are reported in Table \ref{lra_experimental_results}. First, we note that \mname achieves strong results on all tasks consistently compared to the transformer model and significantly outperforms all the other baseline methods and achieve best score in terms of the average accuracy. By taking a closer look at the accuracy for each task, \mname wins over baseline models on four out of five tasks. Notably, \mname can work well on both image and text and the math inference data sets.

\subsection{Pretraining and Finetuning}
\begin{table}[t]
\centering
\begin{tabular}{l|c}
  \toprule \toprule
    \textbf{Setting} &  \textbf{BPC} $\downarrow$\\  \hline
    RoBERTa \citep{liu2019roberta}      & 1.85       \\
    Longformer \citep{beltagy2020longformer}   & 1.71       \\
    BigBird \citep{zaheer2020bigbird}      & 1.68       \\
    \midrule    
    RoBERTa$_{reproduce}$ & 2.05    \\
    \mname & \textbf{1.67}    \\
\bottomrule
\end{tabular}
\caption{MLM BPC for \mname and other models.}
\label{tab:pretrain_bpc}
\end{table}

\subsubsection{Pretraining} In the training task, we follow \citep{liu2019roberta} and pretrain \mname using the Mask Language Model (MLM) training object. This task involves predicting tokens that are randomly masked out. For the training samples, we train \mname with maximum sequence length of 4096 and train it for 1 million steps. Samples with sequence length less than 4096 will be concatenated as one sample to improve training efficiency. For those samples longer than max sequence length, we truncate them to 4096. Statistics of pretraining data can be found in \ref{appendix_pretrain}. Following \citep{beltagy2020longformer}, \mname warm-starts from the public RoBERTa checkpoint and we compare the performance in MLM task in terms of bits per character (BPC). As shown in Table \ref{tab:pretrain_bpc}, BigBird, Longformer, \mname are all better than RoBERTa whose max sequence length is 512. Among those methods, \mname performs best.

\subsubsection{Text Classification }

To test \mname on downstream tasks, we first select three text classification tasks: Arxiv paper categories classification \citep{DBLP:journals/access/HeWLFW19}, IMDB reviews classification \citep{DBLP:conf/acl/MaasDPHNP11} and Hyperpartisan news detection \citep{DBLP:conf/semeval/KieselMSVACSP19}. Arxiv dataset consists of paper text content collected from https://arxiv.org/ and there are 11 classes denoting the paper category. IMDB is a widely used sentiment analysis dataset containing 25,000 movie reviews labeled as positive or negative. Hyperpartisan news dataset contains document for political standpoint classification. To align the experimental settings, we used the dataset split published by \citep{beltagy2020longformer} \footnote{https://github.com/allenai/longformer/blob/master/scripts/hp-splits.json}. 
The experiment was repeated 5 times for both datasets. \mname's hyperparameters are recorded in the appendix \ref{sec:appendix_class}. Note that all linear transformation weights of hierarchical attention are shared with the weights of the previous Sparse Transformer attention. Table \ref{text_cls_experimental_results} summarizes the results of \mname. From this table, it shows that both \mname and BigBird performs better than limited length RoBERTa, with \mname performs the best. For Arxiv, \mname surpasses baseline by a large margin. For IMDB and Hyperpartisan, the performance gain continues demonstrating that \mname is capable of utilizing information from long document input.

\begin{table}[t]
\centering
\resizebox{0.49\textwidth}{!}
{
\begin{tabular}{lccc|c}
  \toprule \toprule
    \textbf{Models} & \textbf{Arxiv}
    \tablefootnote{The reported numbers from \citep{zaheer2020bigbird} on Arxiv dataset are not comparable with ours because they did not release the train/test split of the data. So we re-tested RoBERTa and BigBird's open source checkpoint on our Arxiv train/test split. We open source our Arxiv dataset split: https://github.com/anonymous}
    & \textbf{IMDB}  & \textbf{Hyp} & \multirow{5}{*}{/} \\
    \cline{1-4}
    \#Example         & 30043  &  25000        &   645 & \\
    \#Classed         & 11     &   2           &    2  &    \\ \cline{1-4}
    Len. at $P_{50}$  & 14733  &  215          &  516  &     \\
    Len. at $P_{95}$  & 43951  &  748          &  1990 &    \\ \hline\hline
    \textbf{Metric}  & \textbf{F1}     &   \textbf{F1}  &    \textbf{F1} &\textbf{ Avg}   \\ \hline
    RoBERTa           & 86.86  &  95.00        &  87.80 &89.88\\
    Longformer        & 87.65  &  95.49        &  87.19 &90.11\\
    BigBird           & 87.50  &  95.20        &  92.20 &91.63\\
\midrule
    \mname & \textbf{89.05} & \textbf{95.53} & \textbf{92.81} &\textbf{92.46} \\
\bottomrule
\end{tabular}
}
\caption{Performance of various models on development set of benchmark natural language understanding tasks. Len. at $P_k$ means  the $k$-th percentile example length in the corresponding dataset.}
\label{text_cls_experimental_results}
\end{table}


\subsubsection{Question Answering}

\noindent For QA tasks, we choose WikiHop \citep{wikihop} and TriviaQA \citep{triviaqa}. 
WikiHop is a large scale QA dataset which has over 40K samples. It encourages models for text understanding across multiple documents. Specifically, WikiHop provides several contexts and questions and ask the model to select the best answer from candidates. Meanwhile, TriviaQA is a bigger scale QA dataset that contains over 650K question-answer pairs. It requires the model to identify the answer span given a context and a question. The dataset is distantly supervised, which means that there is no golden span, thus we find all identical answers in provided documents. From Table \ref{qa_experimental_results}, we can see that the length at 50 percentile is over 1K and 4K for WikiHop and TriviaQa respectively, denoting that those two datasets consist of very long context.

In \mname model, following \citep{beltagy2020longformer}, we concatenate the answer / question / candidates with special tokens along with the context. The results of WikiHop and TriviaQA are shown in Table \ref{qa_experimental_results}. For those two datasets, we record the reprodueced results of \citep{beltagy2020longformer} and \citep{zaheer2020bigbird} and the score of \mname in the bottom row. From this table, we see that \mname achieves the best results over all those methods. 
\begin{table}[t]

\centering

\resizebox{0.47\textwidth}{!}
{
\begin{tabular}{lccc|c}
  \toprule \toprule
    \textbf{Models}   & \textbf{WikiHop}  & \multicolumn{2}{c|}{\textbf{TriviaQA}} &  \multirow{4}{*}{/} \\ 
    \cline{1-4}
    \#Example         & 43738             & \multicolumn{2}{c|}{110647}              &   \\ \cline{1-4}
    Len. at $P_{50}$  & 1313               & \multicolumn{2}{c|}{4576}               &   \\
    Len. at $P_{95}$  & 3685              & \multicolumn{2}{c|}{5166}                &   \\ \hline \hline
    \textbf{Metric}                 & \textbf{Acc}      & \textbf{F1} & \textbf{EM} & \textbf{Avg}   \\
\hline
    RoBERTa          & 72.40  & 74.30 & -       & -      \\
    Longformer       & 75.18  & 74.53 & 67.00   & 72.23      \\
    ETC              & 73.70  & - & -           & -      \\
    BigBird          & 72.30  & 73.40 & 68.60   & 71.43      \\
\midrule                
    \mname           & \textbf{75.76}  & \textbf{76.47} & \textbf{71.80} & \textbf{74.67} \\
\bottomrule
\end{tabular}
}
\caption{Model comparison for WikiHop and TriviaQA.}
\label{qa_experimental_results}
\end{table}

\subsection{Ablation}
\label{sec:abl}

\noindent \textbf{Effect of pooling method for representative tokens}
In the experiment, we have explored three different pooling methods while keeping other settings unchanged. The results are shown in Table \ref{pooling_abl}. $\operatorname{MEAN}$ and $\operatorname{MAX}$ represent mean pooling and max pooling, respectively. $\operatorname{CLS_g-Only}$ refers to use the first token $\operatorname{CLS}$ only. As presented in Table \ref{pooling_abl}, ${\operatorname{CLS_g-Only}}$ is the worst on average score, and $\operatorname{MEAN}$ performs on par with $\operatorname{MAX}$. Take a closer look at the score each task, it shows that $\operatorname{MAX}$ performs best at both Image and Text, indicating that $\operatorname{MAX}$ is a stable method for predicting strong results and for image tasks, meanwhile the $\operatorname{MEAN}$ can be considered as a candidate for hyperparameters. For $\operatorname{CLS}$, the score drops significantly at both tasks, indicating that the contextual global information is more important for image tasks.

\textbf{Effect of proposed components: HST and SAR} 
Table \ref{abl_comp} shows the performance of \mname by ablating each proposed component. All models are reported with mean results of 5 runs, the hyperparameters keep the same as the official recommendation. From the last column in Table \ref{abl_comp}, we see that the HST improves \mname with 1.0 percent point on average (\#1 vs. \#0). As the default setting is weight not sharing for linear mapping weight in hierarchical attention with the sparse attention layer, we add a group of experiment \#3 to explore the effect of weight sharing on experimental results. We see that with weight sharing, the results drop by 0.27 points on average by (\#3 vs \#2). By comparing \#2 to \#0, we see that SAR is beneficial in modeling the long sequence bringing 1.44 points improvement on average. Except for those observation on average scores, let us take a closer to the how HST and SAR perform in different types of data. Since HST is expected to be more useful on tasks that require global information, let's look at ListOps, Text and Image first. It can be seen that after removing HST, the average decrease is 1-2 points (ListOps, Text and Image in \#1 vs. \#0). Especially, for ListOps, the task need the model to see all characters in order to do valid logical inference, HST have shown its importance (35.97 vs. 37.75). Except for the global information, some tasks require the model has more stability. For example, the image task Pathfinder, dropped 3.5 points after removing SAR (75.25 vs. 78.77) showing the effect of SAR. Except for those component discussed above, the relation between SAR and R-Drop (\#4) \cite{liang2021rdrop} will be discuss later.



\begin{table}[t]
\centering

\begin{tabular}{l|cc|c}
  \toprule \toprule
\multirow{2}{*}{\textbf{Setting}} & \textbf{Image} & \textbf{Text} & \multirow{2}{*}{\textbf{Avg}} \\ \cline{2-3}
                                   & \textbf{Acc}  & \textbf{Acc}      &                      \\ \hline
   MEAN & 45.28 & 63.19 & 54.23 \\
   MAX  & \textbf{46.51} & \textbf{63.45} & \textbf{54.98}  \\
   $\operatorname{CLS_g}$-Only & 42.88 & 61.58 & 52.23 \\
\bottomrule
\end{tabular}

\caption{Ablation for the pooling method of representative tokens in \mname for downstream tasks. }
\label{pooling_abl}
\end{table}

\begin{table*}[th]
    \centering
\begin{center}
\begin{tabular}{l|ccccc|c}
  \hline \hline
    Models &  ListOps & Text  & Retrieval & Image & Pathfinder & Avg \\ 
\hline
    \#0 \mname & 37.75 & 64.47 & 62.64 & 45.28 & 78.77 & 57.60 \\
    \#1 w/o HST & 35.97 & 63.25 & 62.24 & 42.76 & 78.79 & 56.60 \\
    \#2 w/o SAR & 37.36 & 63.54 & 61.87 & 42.77 & 75.25 & 56.16 \\
    \#3 w/o SAR + ws in HST & 37.41 & 62.55 & 60.32 & 42.62 & 76.55 & 55.89 \\    
    \#4 w/o SAR + R-Drop & 37.68 & 61.73 & 60.59 & 43.48 & 78.11 & 56.32 \\
    
\hline \hline
\end{tabular}
\end{center}
\setlength{\abovecaptionskip}{-0.8mm}

\caption{Performance of \mname by ablating each proposed components. \textbf{ws} means the linear  weights for hierarchical and sparse attention are shared.
}
\label{abl_comp}
\end{table*}

\begin{figure*}[ht]
     \centering
     \begin{subfigure}[t]{0.19\textwidth}
         \centering
         \includegraphics[width=\textwidth]{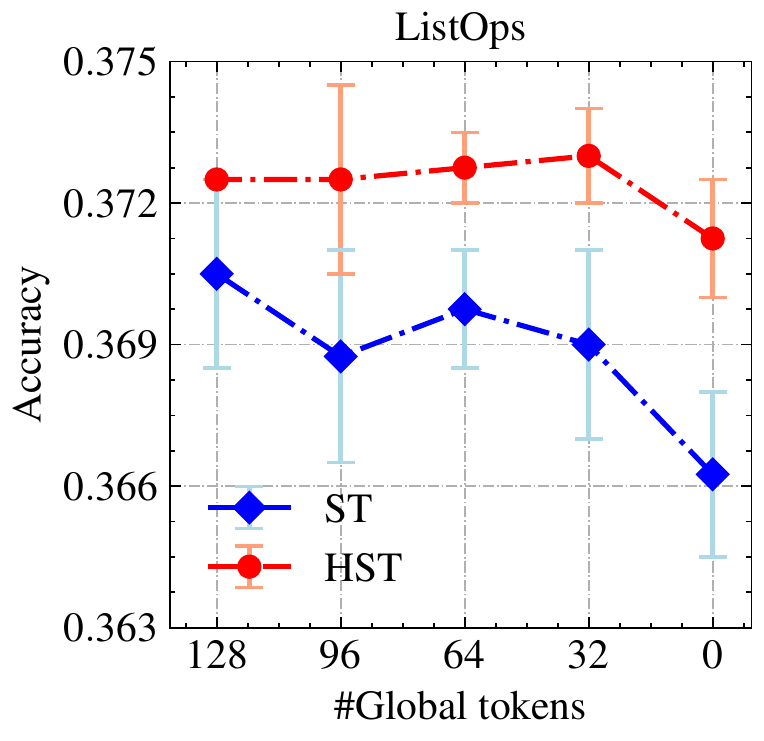}
         \label{fig:listops}
     \end{subfigure}
     \hfill
     \begin{subfigure}[t]{0.19\textwidth}
         \centering
         \includegraphics[width=\textwidth]{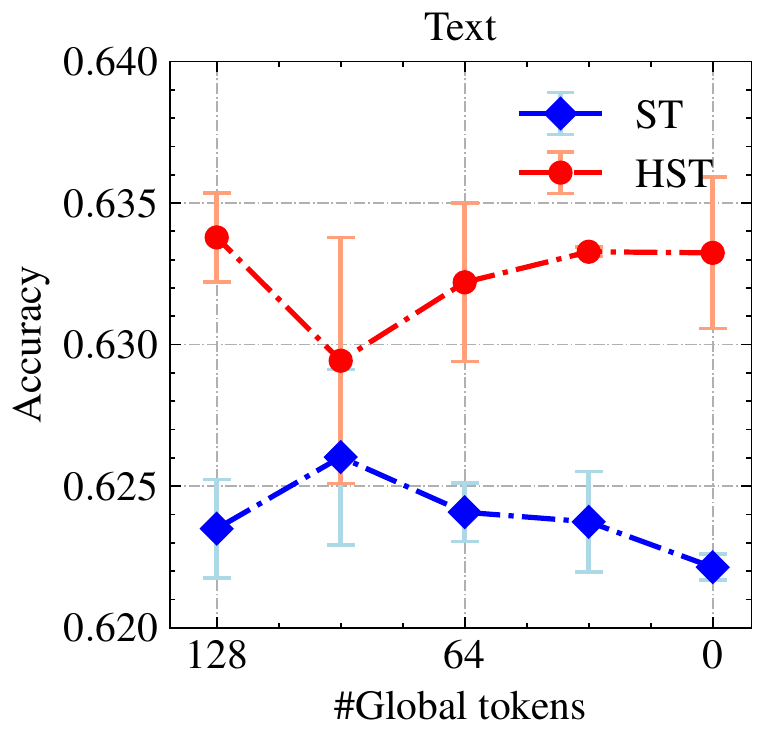}
         \label{fig:text}
     \end{subfigure}
     \hfill
     \begin{subfigure}[t]{0.19\textwidth}
         \centering
         \includegraphics[width=\textwidth]{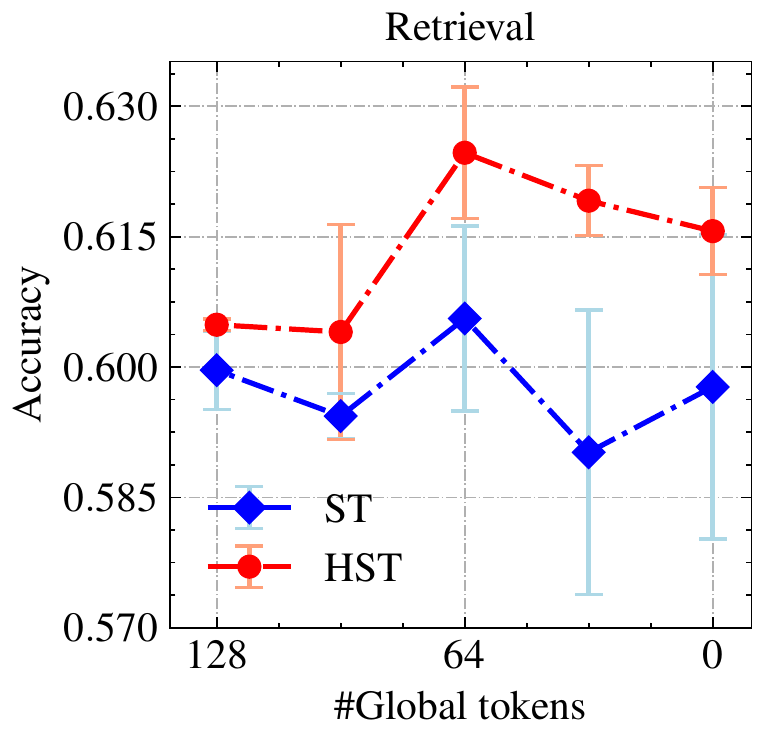}
         \label{fig:retrieval}
     \end{subfigure}
     \hfill
     \begin{subfigure}[t]{0.19\textwidth}
         \centering
         \includegraphics[width=\textwidth]{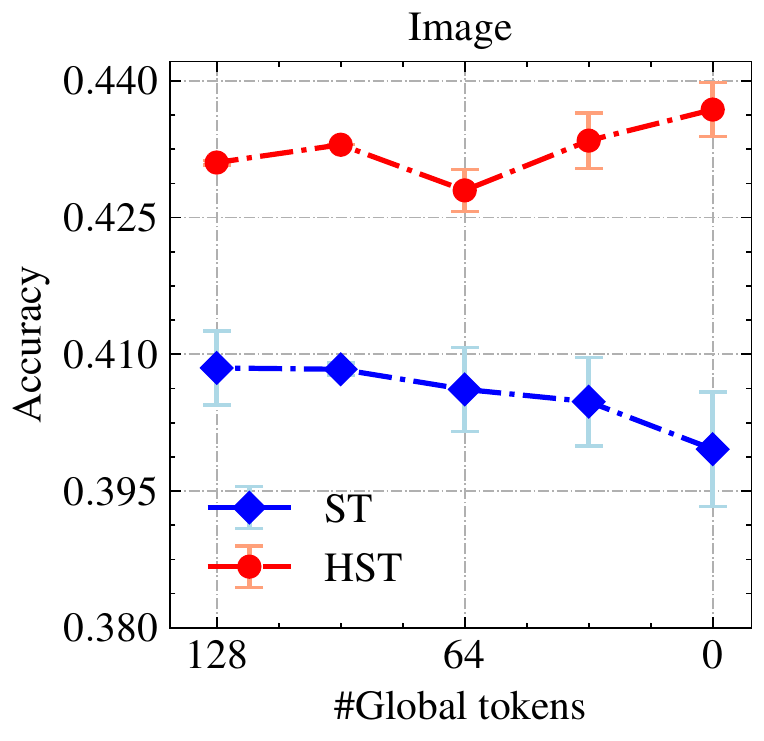}
         \label{fig:image}
     \end{subfigure}
     \hfill
     \begin{subfigure}[t]{0.19\textwidth}
         \centering
         \includegraphics[width=\textwidth]{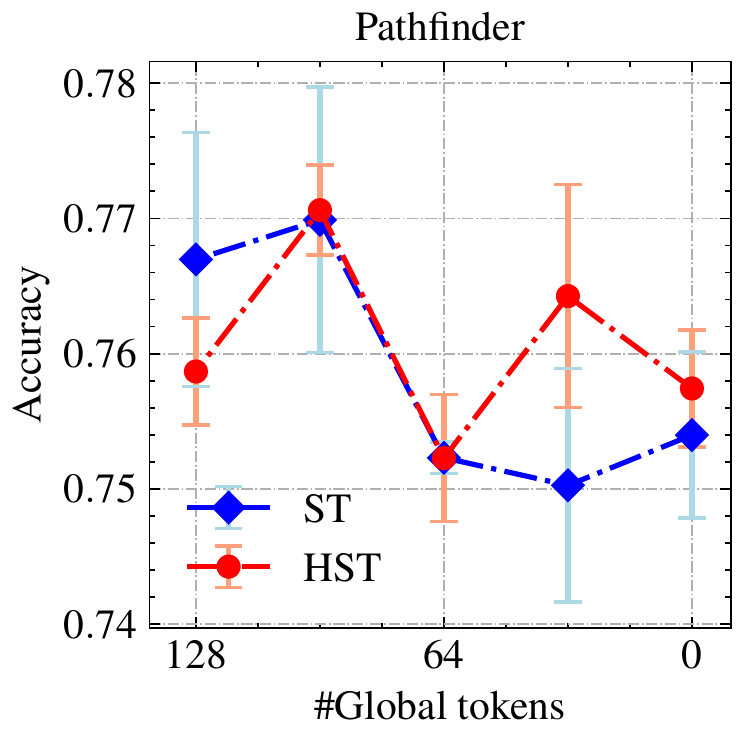}
         \label{fig:pathfinder}
     \end{subfigure}
    \setlength{\abovecaptionskip}{-1mm}
        \caption{Phenomenon of bottleneck and the effectiveness of \textbf{HST}: Accuracy across global token size in the LRA benchmark. A downward trend can be observed for ST on all datasets when bottleneck size decrease (blue line). With our proposed HST method, the performance is less affected by the bottleneck size (red line). Especially, task Text and Image is even better with the \#Global tokens decreasing from 64 to 0. Error bar denotes standard deviation.}
        \label{fig:bottleneck}
\end{figure*}

\noindent \textbf{Effect of SAR vs. R-Drop} In this study, we specifically investigate the importance of Dropout \citep{srivastava2014dropout} in those experiments. As the Dropout \cite{srivastava2014dropout} technique is commonly used in deep neural network training and \citep{liang2021rdrop} use Dropout to constrain the outputs of two subnetworks, we ablate Dropout to compare SAR and R-Drop \citep{liang2021rdrop} as shown by (\#4 vs. \#0) in Table \ref{abl_comp}. \#1 is the SAR-only version \mname result, which means for getting the regularization version model output $\mathcal{P}_{2}\left(y \mid x\right)$ in \ref{section:saor}, we roll the sparse attention input (which is equivalent to shift sparse attention topology) and use the Dropout at the same time, \#4 means that we only use Dropout to get the regularization version model output $\mathcal{P}_{2}\left(y \mid x\right)$. We see that SAR achieves the best for the average score (\#4 vs. \#0). Moreover, for specific task in Table \ref{abl_comp}, our method is more efficient in 3 out of 5, indicating not only SAR's effectiveness, but also that SAR and Dropout are compatible and complementary.

\section{Discussion}
\subsection{Bottleneck Analysis}

\label{section:bottleneck}
In this section, we analyze the impact of bottleneck size on performance of Sparse Transformer and HST respectively. All the hyperparameter configuration follows \citep{tay2020long}. We ran each experiment twice and average the results. The experimental results can be seen in Figure \ref{fig:bottleneck}. As discussed in section \ref{section:revisit}, the number of global tokens determines the size of a bottleneck in Sparse Transformer and we record the trend of Sparse Transformer and HST in LRA dataset by changing the size of bottleneck. In this figure, blue and red line denotes Sparse Transformer and HST respectively. As shown in Figure \ref{fig:bottleneck}, with the information bottleneck size decreases, the performance has a downward trend.

This trend indicates that Sparse Transformer is more prone to the bottleneck size. For our method, the performance of HST in Figure \ref{fig:bottleneck} is less affected by the bottleneck size by using representative token for richer global information interaction. As can be seen in this figure, the task score of HST (red line) in each subfigure can be maintained in the same interval without a downward trend. Especially, task Text and Image is even better with the \#Global tokens decreasing from 64 to 0.

\section{Conclusion}

In this paper, we propose \mname. Firstly, the advantages and disadvantages of Sparse Transformer are analyzed from the perspective of information flow and we introduce HST to provide a richer path for interaction between input tokens. Secondly, we propose a self-attention regularization method, SAR, to improve the consistency of sparse transformers with different attention topologies. Experiments on various downstream tasks demonstrate that \mname outperforms a variety of strong baseline methods including the full-rank attention and other efficient sparse and dense attention methods. 



\clearpage

\bibliography{acl}
\bibliographystyle{acl_natbib}

\appendix
\clearpage

\appendix

\section{Appendix}

\subsection{Pretraining}
\label{appendix_pretrain}
\subsubsection{Pretraining Dataset}

For \mname pretraining, we use Wikipedia (English Wikipedia dump\footnote{https://dumps.wikimedia.org/enwiki/}; 12GB), BookCorpus \citep{bookcorpus} (4.6GB), Realnews \citep{DBLP:conf/nips/ZellersHRBFRC19} ({7.4GB}) and Stories \citep{stories} (11GB). For pretraining, we also sample 5\% training data as the validation set to monitor the training process. Table \ref{pretrain_data} shows statistics of the pretraining data.

\subsubsection{Pretraining Hyperparameters}

We split any document longer than 4096 into multiple documents and we joined multiple documents that were much smaller than 4096. During the pre-training phase, we only use mask language model for training tasks. Specifically, we mask 15\% of tokens in these four datasets, and train \mname to predict the mask. We warm start \mname from RoBERTa’s checkpoint. The hyperparameters for these \mname are given in Table \ref{param_pretrain}. We use a learning rate warmup over the first 10,000 steps, and polynomial decay of the learning rate. Notably, attention weight in HST are shared with sparse attention.

\subsection{Tasks}
To evaluate \mname, we chose three benchmarks, including LRA, and text classification, as well as question answering. The latter two need to follow the pretraining and finetuning paradigm. 
Table \ref{tab:task_statistics} lists the data distribution, task type, evaluation metric of each dataset.

\subsubsection{Hyperparameters for LRA}
Table \ref{tab:hp_lra} gives the detail list of hyperparameters used to get results shown in Table \ref{lra_experimental_results}. 
\label{sec:appendix_class}

\subsubsection{Hyperparameters for Classification and QA}
Table \ref{tab:hp_cls_qa} gives the detail list of hyperparameters used to get results shown in Table \ref{text_cls_experimental_results} and Table \ref{qa_experimental_results}.

\FloatBarrier

\begin{table}[ht]
\centering

\begin{tabular}{l|ccccc|c}
  \toprule \toprule
    Source &  Tokens & Avg doc len  \\ 
\midrule 
    Wikipedia & 3.0B & 515  \\
    BookCorpus & 1.2B & 23K \\
    Realnews & 1.8B & 3.0K   \\
    Stories & 2.7B & 8.7K \\    
\bottomrule
\end{tabular}
\caption{Pretraining data statistics.}
\label{pretrain_data}
\end{table}
\FloatBarrier

\FloatBarrier
\begin{table}[h]
\centering

\begin{tabular}{l|c}
  \toprule \toprule
  \textbf{Parameter} & \textbf{\mname} \\ 
\midrule 
 $\alpha$ of $L_{SAR}$ &  0            \\
 learning rate       &     3$e{-}5$     \\    
 batch size          &     256          \\       
 weight decay        &     0.1          \\       
 warmup steps        &     10k          \\       
 total steps         &     1m          \\       
 max seq length      &     4096         \\       
 embedding dim       &     768          \\       
 \#head              &     12           \\       
 \#layer             &     12           \\       
 activation layer    &     gelu         \\
 dropout             &     0.1          \\       
 attn dropout        &     0.1          \\  
\bottomrule
\end{tabular}
\caption{Hyperparameters for the \mname for Pretraining.}
\label{param_pretrain}
\end{table}

\FloatBarrier

\begin{table*}[t]
\small
\begin{center}
\begin{tabular}{lccccc}
  \toprule \toprule
    \textbf{Hyperparameter} & \textbf{Arxiv} & \textbf{IMDB} & \textbf{Hyperpartisan} & \textbf{WikiHop} & \textbf{TriviaQA}\\ 
\midrule 
    HST pooling        & mean  & mean & mean & mean & mean \\
    $\alpha$ of $\mathcal{L}_{ SAR }$     & 10    & 0.1  & 0    & 10   & 3    \\
    \#roll tokens of $\mathcal{L}_{ SAR }$   & 8     & 8    & 0    & 8    & 8    \\
    learning rate      & 6e-5  & 1e-5 & 3e-5 & 3e-5 & 3e-5 \\ 
    batch size         & 48    & 64   & 16   & 48   & 32   \\
    epoch              & 10    & 20   & 20   & 30   & 10   \\
    warmup             & 10\%  & 10\% & 10\% & 200 (steps)  & 10\% \\
    max seq len        & 4096  & 2048 & 1024 & 4096 & 4096 \\
    \#global token     & \multicolumn{5}{c}{128}   \\
    local window size  & \multicolumn{5}{c}{192}   \\
    \#random token     & \multicolumn{5}{c}{192}   \\
    Optimizer          & \multicolumn{5}{c}{Adam}  \\
    Weight Sharing Option & \multicolumn{5}{c}{ON}  \\
\bottomrule
\end{tabular}
\end{center}
\caption{Hyperparameters of classification and question answering tasks. }
\label{tab:hp_cls_qa}
\end{table*}
\begin{table*}[t]
\small
\begin{center}
\begin{tabular}{lccccc}
  \toprule \toprule
    \textbf{Hyperparameter} & \textbf{ListOps} & \textbf{Text} & \textbf{Retrieval} & \textbf{Image} & \textbf{Pathfinder}\\ 
\midrule
    \multicolumn{6}{l}{\textbf{Hyperparameters for HST and SAR}} \\
\midrule 
    HST pooling         &  \multicolumn{5}{c}{\{ $\operatorname{MIN}, \operatorname{MEAN}, \operatorname{MAX}$\}} \\
    $\alpha$ of $\mathcal{L}_{ SAR }$   & \multicolumn{5}{c}{\{ $0.5, 5, 10$\}}   \\
    \#roll tokens of $\mathcal{L}_{ SAR }$   & \multicolumn{5}{c}{ \{ ${2,8}$ \} }    \\
    Weight Sharing Option  & \multicolumn{5}{c}{OFF} \\
\midrule
    \multicolumn{6}{l}{\textbf{Fixed hyperparameters provided by LRA} \citep{lra}} \\
\midrule 
    learning rate       & 5e-2  & 5e-2  & 5e-2  & 5e-4  & 1e-3  \\
    batch size          & 32    & 32    & 32    & 256   & 512   \\
    weight decay        & 1e-1  & 1e-1  & 1e-1  & 0     & 0     \\
    warmup              & 1000  & 8000  & 8000  & 175   & 312   \\
    max seq length      & 2000  & 1000  & 4000  & 1024  & 1024  \\
    embedding dim        & 512   & 256   & 128   & 32    & 64    \\
    \#head               & 8     & 4     & 4     & 1     & 2     \\
    \#layer              & 4     & 4     & 4     & 1     & 4     \\
    Q/K/V dim           & 512   & 256   & 128   & 32    & 32    \\
    MLP dim             & 1024  & 1024  & 512   & 64    & 64    \\
    dropout             & 0.1   & 0.1   & 0.1   & 0.3   & 0.2   \\
    attn dropout        & 0.1   & 0.1   & 0.1   & 0.2   & 0.1   \\
    lr decay            & root square & root square & root square & cosine & cosine \\
\bottomrule
\end{tabular}
\end{center}
\caption{
The upper part is the hyperparameter related to \mname, while the lower part is the fixed hyperparameter provided by LRA and cannot be changed.}
\label{tab:hp_lra}
\end{table*}

\begin{table*}[t]
\begin{center}
\resizebox{0.98\textwidth}{!}
{
\begin{tabular}{l|c|c|c|c|c|c|c|c|c}
\toprule \toprule
\multirow{2}{*}{\textbf{Corpus}} & \multirow{2}{*}{\textbf{Task}}            & \multirow{2}{*}{\textbf{Split}} & \multirow{2}{*}{\textbf{\#Sample}} & \multicolumn{4}{c|}{\textbf{Length in percentile}}           & \multirow{2}{*}{\textbf{\#Label}} & \multirow{2}{*}{\textbf{Metrics}} \\ \cline{5-8}
                                 &                                           &                                 &                                    & \textbf{50\%} & \textbf{90\%} & \textbf{95\%} & \textbf{max} &                                   &                                   \\ \hline
\multicolumn{10}{c}{\textbf{Long Range Arena (LRA)}}                                                                                                                                                                                                                                     \\ \hline
\multirow{3}{*}{ListOps}         & \multirow{3}{*}{Logical Reasoning}        & Train                           & 96k                                & 954           & 1646          & 1800          & 1999         & \multirow{3}{*}{10}               & \multirow{3}{*}{Acc}              \\ \cline{3-8} 
                                 &                                           & Dev                             & 2k                                 & 960           & 1648          & 1813          & 1999         &                                   &                                   \\ \cline{3-8} 
                                 &                                           & Test                            & 2k                                 & 947           & 1657          & 1803          & 1999         &                                   &                                   \\ \hline
\multirow{2}{*}{Text}            & \multirow{2}{*}{Sentiment Classification} & Train                           & 25k                                & 979           & 2615          & 3431          & 13704        & \multirow{2}{*}{2}                & \multirow{2}{*}{Acc}              \\ \cline{3-8}
                                 &                                           & Dev                             & 25k                                & 962           & 2543          & 3333          & 12988        &                                   &                                   \\ \hline
\multirow{3}{*}{Retrieval}       & \multirow{3}{*}{Retrieval}                & Train                           & 147k                               & 7648          & 13467         & 20495         & 72885        & \multirow{3}{*}{2}                & \multirow{3}{*}{Acc}              \\ \cline{3-8}
                                 &                                           & Dev                             & 18k                                & 7665          & 13359         & 19928         & 72885        &                                   &                                   \\ \cline{3-8}
                                 &                                           & Test                            & 17k                                & 7702          & 15955         & 22427         & 50012        &                                   &                                   \\ \hline
\multirow{3}{*}{Image}           & \multirow{3}{*}{Category Classification}  & Train                           & 45k                                & -             & -             & -             & 1024         & \multirow{3}{*}{10}               & \multirow{3}{*}{Acc}              \\ \cline{3-8} 
                                 &                                           & Dev                             & 5k                                 & -             & -             & -             & 1024         &                                   &                                   \\ \cline{3-8} 
                                 &                                           & Test                            & 10k                                & -             & -             & -             & 1024         &                                   &                                   \\ \hline
\multirow{3}{*}{PathFinder}      & \multirow{3}{*}{Image Reasoning}          & Train                           & 160k                               & -             & -             & -             & 1024         & \multirow{3}{*}{2}                & \multirow{3}{*}{Acc}              \\ \cline{3-8}
                                 &                                           & Dev                             & 20k                                & -             & -             & -             & 1024         &                                   &                                   \\ \cline{3-8}
                                 &                                           & Test                            & 20k                                & -             & -             & -             & 1024         &                                   &                                   \\ \hline
\multicolumn{10}{c}{\textbf{Text Classification}}                                                                                                                                                                                                                                        \\ \hline
\multirow{2}{*}{Arxiv}           & \multirow{2}{*}{Category Classification}  & Train                           & 33k                                & 14733         & 34209         & 43951         & 1121751      & \multirow{2}{*}{11}               & \multirow{2}{*}{Micro F1}         \\ \cline{3-8}
                                 &                                           & Test                            & 3.3k                               & 14710         & 32417         & 40965         & 850540       &                                   &                                   \\ \hline
\multirow{2}{*}{IMDB}            & \multirow{2}{*}{Sentiment Classification} & Train                           & 25k                                & 215           & 569           & 748           & 3084         & \multirow{2}{*}{2}                & \multirow{2}{*}{Micro F1}         \\ \cline{3-8}
                                 &                                           & Test                            & 25k                                & 212           & 550           & 724           & 2778         &                                   &                                   \\ \hline
\multirow{3}{*}{Hyperpartisan}   & \multirow{3}{*}{News Classification}      & Train                           & 516                                & 536           & 1517          & 1990          & 5560         & \multirow{3}{*}{2}                & \multirow{3}{*}{Micro F1}         \\ \cline{3-8}
                                 &                                           & Dev                             & 65                                 & 520           & 1535          & 1971          & 2637         &                                   &                                   \\ \cline{3-8}
                                 &                                           & Test                            & 65                                 & 637           & 1771          & 1990          & 5560         &                                   &                                   \\ \hline
\multicolumn{10}{c}{\textbf{Question Answering}}                                                                                                                                                                                                                                         \\ \hline
\multirow{2}{*}{TriviaQA}        & \multirow{2}{*}{Question Answering}       & Train                           & 110k                               & 4576          & 5027          & 5166          & 10091       & \multirow{2}{*}{Span}             & \multirow{2}{*}{Macro F1 \& EM}   \\ \cline{3-8}
                                 &                                           & Dev                             & 14k                                & 4577          & 5026          & 5169          & 10210       &                                   &                                   \\ \hline
\multirow{2}{*}{WikiHop}         & \multirow{2}{*}{Question Answering}       & Train                           & 43k                                & 1313          & 3001          & 3685          & 19747        & \multirow{2}{*}{Candidates}       & \multirow{2}{*}{Acc}              \\ \cline{3-8}
                                 &                                           & Dev                             & 5.1k                               & 1413          & 3184          & 3871          & 17004        &                                   &                                   \\ \bottomrule
\end{tabular}
}
\end{center}
\caption{ Downstream tasks statistics. Samples of tasks Image and PathFinder are all $32\times32$ images. }
\label{tab:task_statistics}
\end{table*}

\subsection{Complexity Analysis}
\label{sec:complexity_analysis}
   


In this section, we analyze the complexity of \mname. Note that the two techniques of \mname, HST and SAR, are general and can be applied with any type of sparse patterns. Considering that only HST of the two techniques affects computation complexity, we will only analyze the performance of HST here. In order to analyze the complexity, we will follow the following process: firstly, we select a sparse attention and secondly, we analyze the HST complexity to compare the additional computation complexity brought by HST. For the sparse pattern we choose to use the recently proposed pattern from BigBird~\citep{zaheer2020bigbird} for comparison, and the computation complexity is $O(n)$. For HST layer, as it is a dense attention for $m$ representation tokens, the computation complexity is $O(m^2)$. Though the computation complexity for HST is quadratic, this will not bring over computation overhead as the number of representative tokens $m\ll n$ in practice. 



\subsection{More details for SAR}

\paragraph{Training with doubled steps.} As discussed in Section \ref{section:saor}, we use the SAR as a regularization term in the total loss. To achieve this training goal, we firstly halve the batchsize and then copy the batch samples to keep the total batchsize is unchanged. In order to ensure sufficient convergence of SAR, we double the training steps. Therefore, it is necessary to see the effect for changing training steps. 

\begin{figure*}
\centering
\begin{subfigure}[b]{.75\textwidth}
\includegraphics[width=\textwidth]{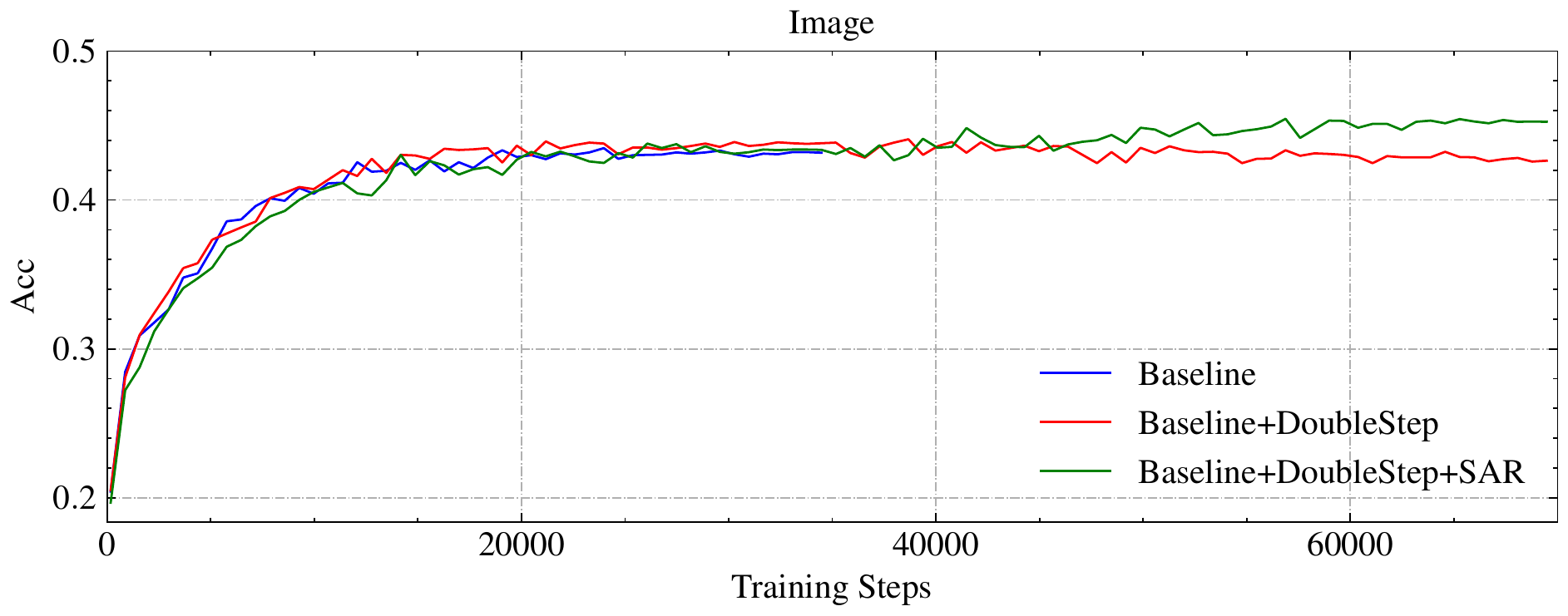}%
\end{subfigure}\hfill
\begin{subfigure}[b]{.75\textwidth}
\includegraphics[width=\textwidth]{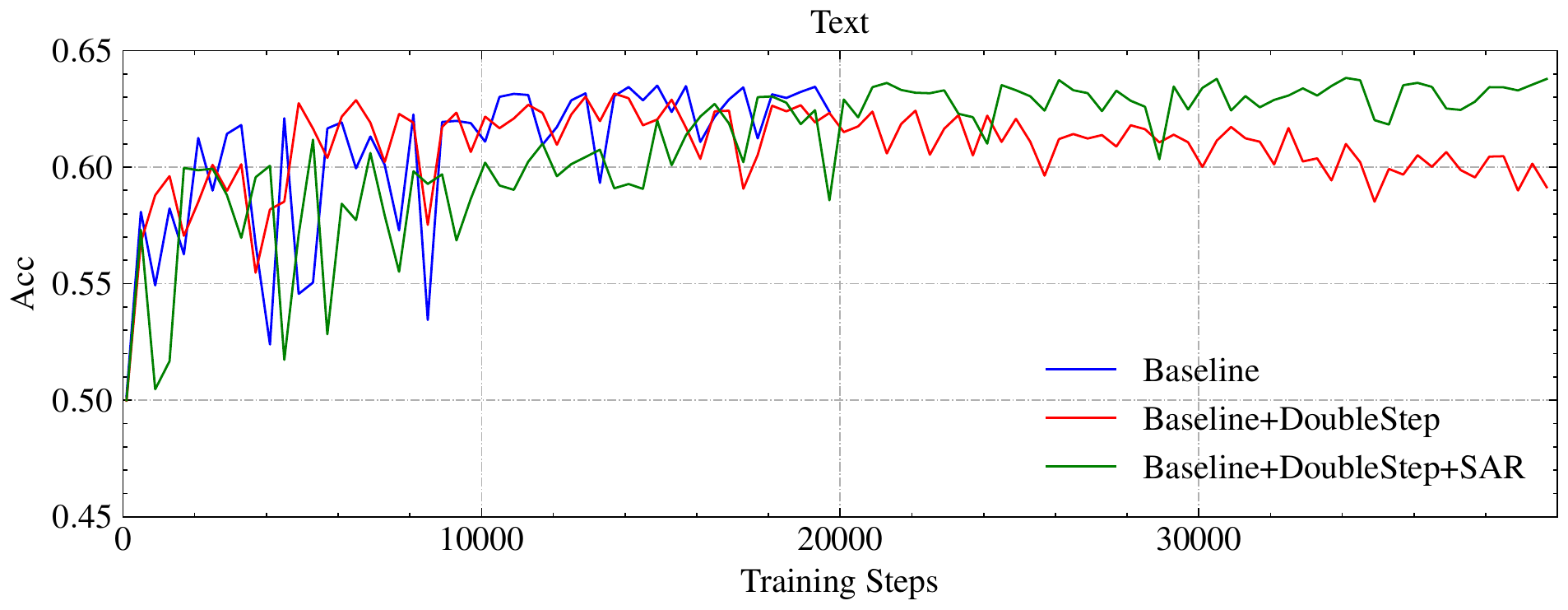}%
\end{subfigure}%
\caption{Results of SAR and Baseline on task Image and Text with doubled training steps.}
\label{soar_overhead}
\end{figure*}

To conduct those experiments, we choose Image and Text in LRA as the ablation dataset. Using HST as the baseline model which is denoted as $\operatorname{Baseline}$ in Figure \ref{soar_overhead}, and we double the training steps for HST which is denoted as $\operatorname{Baseline+DoubleStep}$. Finally, we apply SAR on $\operatorname{Baseline}$. All of those results are recorded in Figure \ref{soar_overhead}. We can see that $\operatorname{Baseline+DoubleStep}$ shows the trend for overfitting, while $\operatorname{Baseline+DoubleStep+SAR}$ has a better convergence curve. For the additional training cost of SAR, we record the training speed for Text. For the $\operatorname{Baseline}$, the speed is 59 samples per second, and 56 samples per second after applying SAR. The additional cost is from the KL-divergence loss backward computation. We can see the cost is 1.03 times, which is a negligible cost. 

\paragraph{How to Shift the sparse attention topology.} SAR is devised upon self-attention and requires different attention topologies for same model input $x$, which is defined in \ref{section:revisit}. Actually, to get a shifted version attention topology, a quick implementation is that we can roll the attention input $H$, as shown in Figure \ref{fig:saor}. Furthermore, we can choose any layer for applying the rolling operator to get the shifted version attention topology. However, note that if the SAR is applied from the first layer, the rolling operator should be applied after the word embedding adding the position embedding. 


\end{document}